\documentclass[11pt]{article}
\usepackage{acl2012}
\usepackage{times}
\usepackage{latexsym}
\usepackage{amsmath}
\usepackage{amssymb}
\usepackage{multirow}
\usepackage{url}
\usepackage{mathtools}
\usepackage{pdfpages}

\usepackage[utf8]{inputenc}
\usepackage{amsmath}
\usepackage{graphicx}
\usepackage{qtree}
\usepackage{tikz}
\usepackage{verbatim}
\usepackage{amsmath}
\usepackage[noend]{algpseudocode}
\usepackage{algorithm}
\usepackage[T1]{fontenc}
\usepackage{pgfplots}
\usepackage{tikz-dependency}
\usepackage{adjustbox}
\usepackage{tabularx}
\usepackage{enumitem}
\setitemize{noitemsep,topsep=0pt,parsep=0pt,partopsep=0pt}

\makeatletter
\renewcommand{\Function}[2]{%
  \csname ALG@cmd@\ALG@L @Function\endcsname{#1}{#2}%
  \def\jayden@currentfunction{#1}%
}
\newcommand{\funclabel}[1]{%
  \@bsphack
  \protected@write\@auxout{}{%
    \string\newlabel{#1}{{\jayden@currentfunction}{\thepage}}%
  }%
  \@esphack
}
\makeatother

\algrenewcommand\algorithmicindent{1.0em}%

\title{Easy-First Dependency Parsing with Hierarchical Tree LSTMs}

\author{Eliyahu Kiperwasser \\
  Computer Science Department \\
  Bar-Ilan University \\
  Ramat-Gan, Israel \\
  {\tt elikip@gmail.com} \\\And
  Yoav Goldberg \\
  Computer Science Department \\
  Bar-Ilan University \\
  Ramat-Gan, Israel \\
  {\tt yoav.goldberg@gmail.com} \\}

\date{}

\begin{document}
\maketitle
\begin{abstract}

  We suggest a compositional vector representation of parse trees that
  relies on a recursive combination of recurrent-neural network encoders.
  To demonstrate its effectiveness, we use the representation as the backbone of
  a greedy, bottom-up dependency parser, achieving very strong accuracies
  for English and Chinese, without relying on external word embeddings. 
  The parser's implementation is available for download at the first author's
  webpage.

\end{abstract}
\section{Introduction}
Dependency-based syntactic representations of sentences are central to many
language processing tasks \cite{kubler2008dependency}. Dependency parse-trees encode not only the
syntactic structure of a sentence but also many aspects of its semantics.

A recent trend in NLP is concerned with encoding sentences as a
vectors (``sentence embeddings''), which can then be used for further prediction
tasks.  Recurrent neural networks (RNNs) \cite{elman1990finding}, and in particular methods based on the
LSTM architecture \cite{hochreiter1997long}, work very well for modeling sequences, and
constantly obtain state-of-the-art results on both language-modeling and
prediction tasks (see, e.g. \cite{mikolov2010recurrent}).

Several works attempt to extend recurrent neural networks to work on trees (see
Section \ref{sec:related} for a brief overview), giving rise to the so-called
recursive neural networks \cite{goller1996learning,socher2010learning}. However, recursive neural networks do not cope well
with trees with arbitrary branching factors -- most work require the encoded trees
to be binary-branching, or have a fixed maximum arity.  Other attempts allow
arbitrary branching factors, in the expense of ignoring the order of the
modifiers.

In contrast,
we propose a tree-encoding that naturally supports trees with arbitrary branching factors, making it particularly
appealing for dependency trees.  Our tree encoder uses recurrent neural networks
as a building block: we model the left and right sequences of modifiers using
RNNs, which are composed in a recursive manner to form a tree (Section
\ref{sec:tree}).
We use our tree representation for encoding the partially-build parse trees in a
greedy, bottom-up dependency parser which is based on the easy-first
transition-system of Goldberg and Elhadad \shortcite{goldberg2010efficient}.  

Using the Hierarchical Tree LSTM representation, and without using any external
embeddings, our parser achieves parsing
accuracies of 92.6 UAS and 90.2 LAS on the PTB (Stanford dependencies) and 
86.1 UAS and 84.4 LAS on the Chinese treebank, while relying on greedy decoding.

To the best of our knowledge, this is the first work to demonstrate competitive
parsing accuracies for full-scale parsing while relying solely on recursive,
compositional tree representations, and without using a reranking framework.
We discuss related work
in Section \ref{sec:related}.

While the parsing experiments demonstrate the suitability of our representation
for capturing the structural elements in the parse tree that are useful for
predicting parsing decisions, we are interested in exploring the use of the RNN-based
compositional vector representation of parse trees also for semantic tasks such
as sentiment analysis \cite{socher2013recursive,tai2015improved}, sentence similarity judgements \cite{sick}
and textual entailment \cite{snli}.

\section{Background and Notation}

\subsection{Dependency-based Representation}

A dependency-based syntactic representation is centered around syntactic modification
relations between head words and modifier words. The result are trees in which
each node is a word in the sentence, and every node except for one designated
root node has a parent node. A dependency tree over a sentence with $n$ words
$w_1,\dots,w_n$ can be represented as a list of $n$ pairs of the form $(h,m)$,
where $0 \leq h \leq n$ and $1 \leq m \leq n$. Each such pair represents an edge
in the tree in which $h$ is the index of a head word
(including the special ROOT node 0), and $m$ is the index of a modifier word.
In order for the dependency trees to be useful for actual downstream language
processing tasks, each edge is labeled with a
syntactic relation.  The tree representation then becomes a list of triplets
$(h,m,\ell)$, where $1 \leq \ell \leq L$ is the index of a dependency relation out of a designated
set of $L$ syntactic relations.

Dependency trees tend to be relatively shallow, with some nodes having many
children.  
Looking at trees in the PTB training set we find that 94\% of the trees have a
height 
of at most
10, and 49\% of the trees a height of at most 6.  In terms of width, 93\% of the
trees have at least one node with an arity of 4 or more, and 56\% of the trees have at 
least one node with an arity of 6 or more.

\subsection{Recurrent Networks and LSTMs}

Recurrent neural networks (RNNs), first proposed by Elman \shortcite{elman1990finding} are statistical
learners for modeling sequential data.  In this work, we use the RNN abstraction as a building
block, and recursively combine several RNNs to obtain our tree representation.
We briefly describe the RNN abstraction below. For further detail on RNNs,
the reader is referred to sources such as
\cite{goldberg2015primer,bengio2015deep,cho2015natural}.

The RNN abstraction is a function $RNN$ 
that takes in a sequence of inputs vectors
$x_1,\dots,x_n$ ($x_i \in \mathbb{R}^{d_{in}}$), and produces a sequence of state
vectors
(also called output vectors) $y_1,\dots,y_n$ ($y_i \in \mathbb{R}^{d_{out}}$).
Each $y_i$ is
conditioned on all the inputs $x_1,\dots,x_i$ preceding it.
Ignoring the intermediate outputs $y_1,\dots,y_{n-1}$, the RNN can be thought of
as encoding the sequence $x_1,\dots,x_n$ into a final state $y_n$. 
Our notation in this paper follows this view.

The RNN is defined recursively using two functions:\footnote{We follow the
notation of Goldberg \shortcite{goldberg2015primer}, with the exception of taking the output
of the RNN to be a single vector rather than a sequence, and renaming $R$ to
$N$.}\\

\noindent $    RNN(s_0,x_{1},\ldots,x_n) = y_n = O(s_n) $\\
$ s_i = N(s_{i-1}, x_i) $\\

Here, a function $N$ takes as input a vector $x_i$ and a state vector
$s_{i-1}$ and returns as output a new state $s_i$. One can then extract
an output vector $y_i$ from $s_i$ using the function $O$ (the function
$O$ is usually the identity function, or a function that returns a subset
of the elements in $s_i$).

Taking an algorithmic perspective, one can view
the RNN as a \texttt{state} object with three operations:
\texttt{s = RNN.initial()} returns a new initial state,
\texttt{s.advance(x)} takes an input vector and returns a new state, and
\texttt{s.output()} returns the output vector for the current state.  When clear
from the context, we abbreviate and use the state's name (\texttt{s}) instead of
\texttt{s.output()} to refer to the output vector at the state.

The functions $N$ and $O$ defining the RNN are parameterized by parameters
$\theta$ (matrices and vectors), which are trained from data. Specifically, one is usually interested in using some
of the outputs $y_i$ for making predictions.  The RNN is trained
such that the encoding $y_i$ is good for the prediction task.  That is, the
RNN learns which aspects of the sequence $x_1,\dots,x_i$ are informative for
the prediction.

We use subscripts (i.e. $RNN_L$, $RNN_R$) to indicate different RNNs, that is,
RNNs that have different sets of parameters.

Specific instantiations of $N$ and $O$ yield different recurrent network
mechanisms.  In this work we use the Long Short Term Memory (LSTM) variant 
\cite{hochreiter1997long} which is shown to be a very capable sequence learner.
However, our algorithm and encoding method do not rely on any
specific property of the LSTM architecture, and the LSTM can be transparently 
switched for any other RNN variant.

\section{Tree Representation}
\label{sec:tree}
\vspace{-5pt}

We now describe our method for representing a tree as a $d$-dimensional vector.
We assume trees in which the children are ordered and there are $k_l \geq 0$
children before the parent node (left children) and $k_r \geq 0$ children  after it (right children).
Such trees correspond well to dependency tree structures.
We refer to the parent node as a head, and to its children as modifiers.
For a node $t$, we refer to its left modifiers as $t.l_1,t.l_2,\dots,t.l_{k_l}$
and its right modifiers as $t.r_1,t.r_2,\dots,t.r_{k_r}$
The indices of the modifier are always from the parent outward, that is $t.l_1$ is
the left modifier closest to the head $t$:
\includegraphics[scale=1.0]{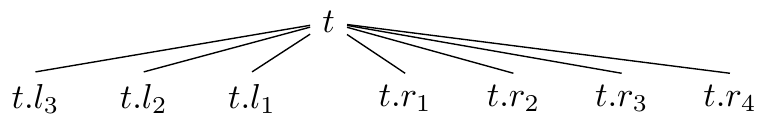}

The gist of the idea is to treat the modifiers of a node as a sequence,
and encode this sequence using an RNN. We separate left-modifiers from right-modifiers,
and use two RNNs: the first RNN encodes the sequence
of left-modifiers from the head outwards, and the second RNN the sequence of
right-modifiers from the head outwards.  The first input to each RNN is the
vector representation of the head word, and the last input is the
vector representation of the left-most or the right-most modifier. The
node's representation is then a concatenation of the RNN encoding of the
left-modifiers with the RNN encoding of the right-modifiers.  The encoding is
recursive: the representation for each of the modifier nodes is computed in a
similar fashion.

\vspace*{0.5em}
\includegraphics[scale=0.8]{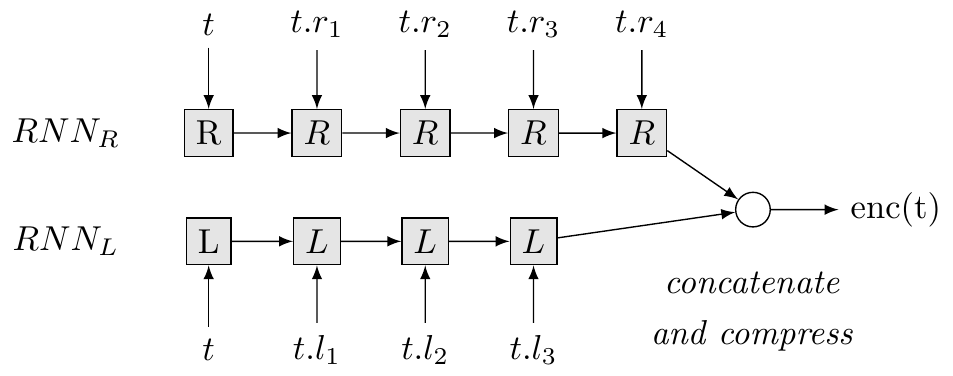}
\vspace*{0.5em}

More formally, consider a node $t$.
Let $i(t)$ be the sentence index of the word
corresponding to the head node $t$, and let $v_i$ be a vector corresponding to
the $i$th word in the sentence (this vector captures information such as the
word form and its part of speech tag, and will be discussed shortly).
The vector encoding of a node $enc(t) \in \mathbb{R}^{d_{enc}}$ is then defined as follows:
\begin{align*}
    enc(t) =& g(W^e \cdot (e_l(t) \circ e_r(t)) + b^e) \\
    e_l(t) =& RNN_L(v_{i(t)}, enc(t.{l_1}), \ldots, enc(t.l_{k_l})) \\ 
    e_r(t) =& RNN_R(v_{i(t)}, enc(t.{r_1}), \ldots, enc(t.r_{k_r})) \\
\end{align*}
First, the sequences consisting of the head-vector $v_{i(t)}$ followed by
left-modifiers and the head-vector followed by right-modifiers are encoded using two RNNs,
$RNN_L$ and $RNN_R$, resulting in RNN states $e_l(t) \in \mathbb{R}^{d_{out}}$
and $e_r(t) \in \mathbb{R}^{d_{out}}$.  Then, the RNN states are concatenated,
resulting in a $2d_{out}$-dimensional vector $(e_l(t) \circ e_r(t))$, which is
reduced back to $d$-dimensions using a linear transformation followed by a non-linear
activation function $g$.
The recursion stops at leaf nodes, for which:
\begin{align*}
    enc(\text{leaf}) =& g(W^e \cdot (e_l(\text{leaf}) \circ e_r(\text{leaf})) + b^e) \\
    e_l(\text{leaf}) =& RNN_L(v_{i(\text{leaf})}) \\ 
    e_r(\text{leaf}) =& RNN_R(v_{i(\text{leaf})}) \\
\end{align*}

\begin{figure*}[t!]
    \begin{center}
    \begin{adjustbox}{width=1\textwidth}
    \includegraphics[scale=1.0]{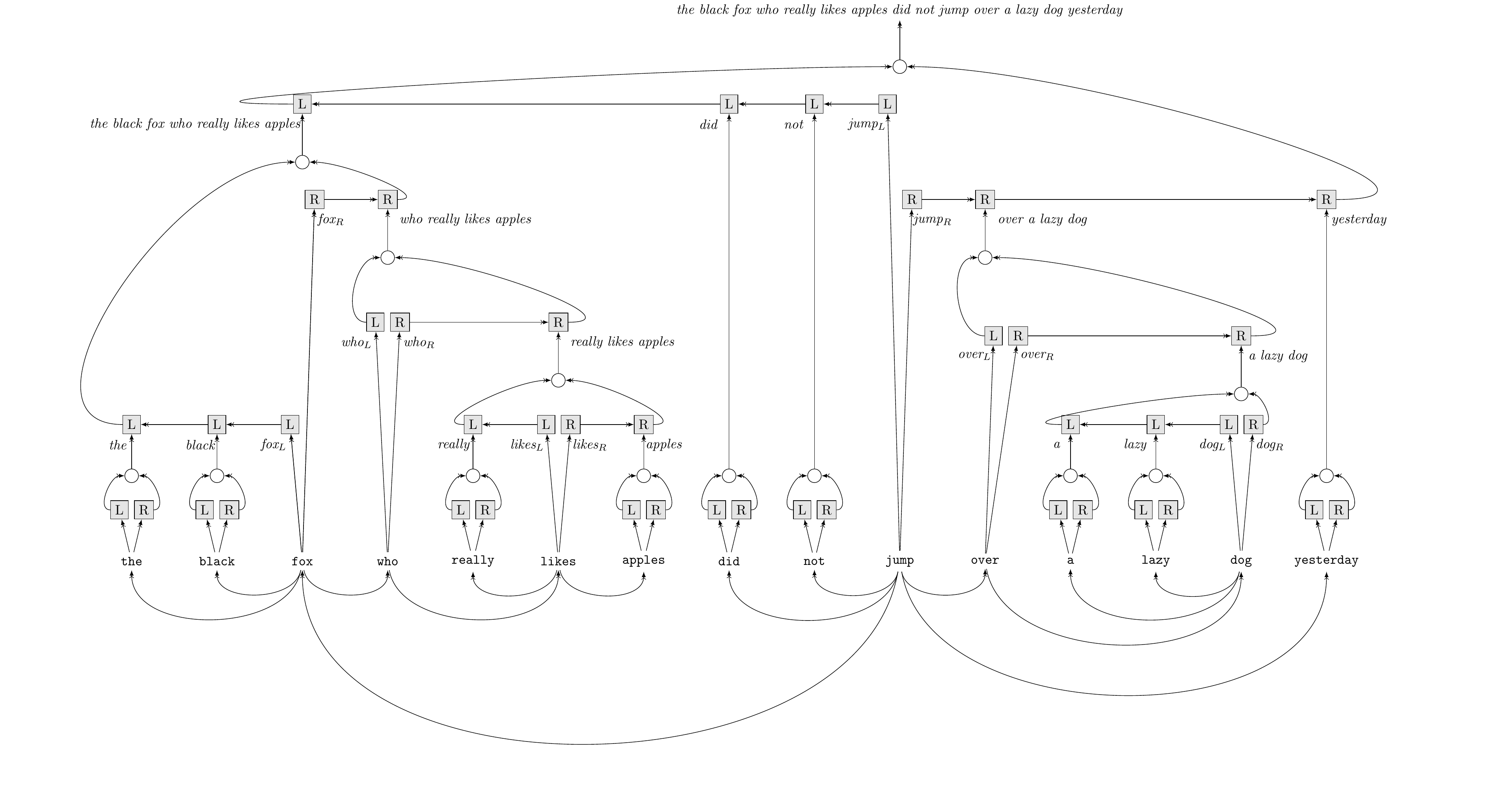}
    \end{adjustbox}
    \end{center}
    \caption{Network for encoding the sentence ``the black fox who really likes
    apples did not jump over a lazy dog yesterday''. \textbf{Top}: the network structure: boxed nodes represent
    LSTM cells, where
    \scalebox{0.7}{\tikz{\node[draw=black,color=black,fill=black!10,shape=rectangle,minimum
    width=0.5em,minimum height=0.5em] {L};}} are cells belonging to the
    left-modifiers sequence model $RNN_L$, and
    \scalebox{0.7}{\tikz{\node[draw=black,color=black,fill=black!10,shape=rectangle,minimum
    width=0.5em,minimum height=0.5em] {R};}} to the right-modifiers sequence
    model $RNN_R$.  Circle nodes represent a concatenation
    followed by a linear transformation and a non-linearity. \textbf{Bottom}:
    the dependency parse of the sentence.}
    \label{fig:example}
    \vspace*{1.0em}
\end{figure*}

Figure \ref{fig:example} shows the network used for encoding the sentence ``the
black fox who really likes apples did not jump over a lazy dog yesterday''. 

\subsection{Representing words}
In the discussion above we assume a vector representation $v_i$ associated with
the $i$th sentence word.  What does $v_i$ look like?  A sensible approach would
be to take $v_i$ to be a function of the word-form and the part-of-speech (POS) tag of the
$i$th word, that is:
\begin{align*}
    v_i = g(W^v \cdot (w_i \circ p_i) + b^v) 
\end{align*}
\noindent where $w_i$ and $p_i$ are the embedded vectors of the word-form
and POS-tag of the $i$th word.

This encodes each word in isolation, disregarding its context. The context of a
word can be very informative regarding its meaning.  One way of incorporating
context is the Bidirectional RNN \cite{schuster1997bidirectional}.
Bidirectional RNNs are shown to be an effective representation for sequence
tagging \cite{irsoy2014opinion}. Bidirectional RNNs represent a word in the sentence using a
concatenation of the end-states of two RNNs, one running from the beginning of
the sentence to the word and the other running from the end to the word. The
result is a vector representation for each word which captures not only the word but also its context.

We adopt the Bidirectional LSTM scheme to enrich our node vector representation,
and for an $n$-words sentence compute the vector representations $v_i$ as
follows:
\begin{align*}
 {v'}_i =& g(W^v \cdot (w_i \circ p_i) + b^v) \\
 f_i =& LSTM_F({v'}_{1}, {v'}_{2}, \dots, {v'}_{i}) \\ 
 b_i =& LSTM_B({v'}_{n}, {v'}_{n-1}, \dots, {v'}_{i}) \\
 v_i =& (f_i \circ b_i)
\end{align*}
We plug this word representation as word vectors, allowing each word vector
$v_i$ to capture information regarding the word form and POS-tag, as well as the
sentential context it appears in.
The BI-LSTM encoder is trained jointly with the rest of the network towards the
parsing objective, using back-propagation.

\paragraph{Embedding vectors}
The word and POS embeddings $w_i$ and $p_i$ are also trained together with the
network.  For the word embeddings, we experiment with random
initialization, as well as with initialization using pre-trained word
embeddings.  
Our main goal in this work is not to provide top parsing accuracies,
but rather to evaluate the ability of the proposed compositional architecture to learn and capture the structural cues that are needed for accurate parsing.
Thus, we are most interested in the random initialization setup:
what can the network learn from the training corpus alone, \emph{without} relying
on external resources.

However, the ability to perform semi-supervised learning by initializing the word-embeddings with vectors
that are pre-trained on large amount of unannotated data is an appealing property of the neural-network approaches,
and we evaluate our parser also in this semi-supervised setup.
When using pre-trained word embeddings, we follow \cite{dyer2015transitionbased}
and use embedding vectors which are trained using \emph{positional context}
\cite{ling2015twotoo},
as these were shown to work better than traditional skip-gram vectors for
syntactic tasks such as part-of-speech tagging and parsing.

\subsection{A note on the head-outward generation}
Why did we choose to encode the children from the head outward, and not the
other way around?  The head outward generation order is needed to
facilitate incremental tree construction and allowing for efficient parsing,
as we show in section \ref{sec:parsing} below.  Beside the efficiency
considerations, using the head-outward encoding puts more emphasis on the
outermost dependants, which are known to be the most informative for predicting
parse structure.\footnote{Features in transition-based dependency parsers often
look at the current left-most and right-most dependents of a given node,
and almost never look further than the second left-most or second right-most
dependents.  Second-order graph based dependency parsers
\cite{mcdonald-phd,eisner2000bilexical}
also condition on the current outermost dependent when generating its sibling.}
We rely on the RNN capability of extracting information from arbitrary positions
in the sequence to incorporate information about the head word itself, which
appears in the beginning of the sequence.  This seem to work well, which is
expected considering that the average maximal number of siblings in one
direction in the PTB is 4.1, and LSTMs were demonstrated to capture much longer-range interactions.
Still, when using the tree encoding in a situation where the tree is fully
specified in advance, i.e. for sentence classification, sentence similarity or
translation tasks, using a head-inward generation order (or even a
bi-directional RNN) may prove to work better.  We leave this line of inquiry to
future work.

The head-outward modifier generation approach has a long history in the parsing
literature, and goes back to at least Eisner \shortcite{eisner1996three} and
Collins \shortcite{collins1997three}.
In contrast to previous work in which each modifier could condition only on a
fixed small number of modifiers preceding it, and in which the left- and right- sequences of
modifiers were treated as independent from one another for computational
efficiency reasons, our approach allows the model to access information from the
entirety of both the left and the right sequences jointly.

\vspace{-3pt}
\section{Parsing Algorithm}
\label{sec:parsing}
\vspace{-3pt}

We now turn to explain how to parse using the tree encoder defined above.  We
begin by describing our bottom-up parsing algorithm, and then show how the
encoded vector representation can be built and maintained throughout the parsing
process.

\subsection{Bottom-up Parsing}

We follow a (projective) bottom-up parsing strategy, similar to the easy-first parsing
algorithm of Goldberg and Elhadad \shortcite{goldberg2010efficient}.

The main data-structure in the parser is a list of partially-built parse trees
we call \emph{pending}.  For a sentence with words $w_1,\dots,w_n$, the \emph{pending} list is initialized with
$n$ nodes, where $pending[i]$ corresponds to word $w_i$.  The algorithm then
chooses two neighbouring trees in the \emph{pending} list $pending[i]$ and
$pending[i+1]$ and either attaches the root of $pending[i+1]$ as the right-most modifier of
the root of $pending[i]$, or attaches the root of $pending[i]$ as the left-most
modifier of the root of $pending[i+1]$.  The tree which was treated as modifier
is then removed from the $pending$ list, shortening it by one.  The process ends
after $n-1$ steps, at which we remain with a single tree in the pending list,
which is taken to be the output parse tree. 
The parsing process is described in Algorithm \ref{alg:parsing}.

\begin{algorithm}[th]
\begin{algorithmic}[1]

    \State{\textbf{Input:} Sentence $w = w_1,\dots,w_n$}
    \For{$i \in 1,\dots,n$}
        \State{pend$[i].id \gets i$}
    \EndFor
    \State{arcs $\gets []$}

    \While{$|\text{pend}|$ > 1}
    \State{$A \gets \{ (i,d) \mid 1 \leq i < |\text{pend}|, d \in \{l,r\} \}$ }
    \State{$i,d \gets select(A)$} \label{line:select}
    \If{$d = l$} 
        \State{$m,h \gets \text{pend}[i],\text{pend}[i+1]$}
        \State{pend.remove($i$)}
    \Else
        \State{$h,m \gets \text{pend}[i],\text{pend}[i+1]$}
        \State{pend.remove($i+1$)}
    \EndIf
    \State{arcs.append($h.id,m.id$)}
    \EndWhile{}
    \State{\Return arcs}
\end{algorithmic}
\caption{Parsing}\label{alg:parsing}
\end{algorithm}

This parsing algorithm is both sound and complete with respect to the class of
projective dependency trees \cite{goldberg2010efficient}. The algorithm depends on non-deterministic choices
of an index in the pending list and an attachment direction (line
\ref{line:select}).
When parsing in practice, the
non-deterministic choice will be replaced by using a trained classifier to assign
a score to each index-direction pair, and selecting the highest scoring pair.
We discuss the scoring function in Section \ref{sec:scoring}, and the training
algorithm in Section \ref{sec:training}.

\subsection{Bottom-up Tree-Encoding}
We would like the scoring function to condition on the vector
encodings of the subtrees it aims to connect.  Algorithm \ref{alg:parsing2}
shows how to maintain the
vector encodings together with the parsing algorithm, so that at every stage in
the parsing process each item $pending[i]$ is associated with a vector encoding
of the corresponding tree.

\begin{algorithm}[th]
\begin{algorithmic}[1]

    \State{\textbf{Input:} Sentence $w = w_1,\dots,w_n$}
    \State{\textbf{Input:} Vectors $v_i$ corresponding to words $w_i$}
    \State{arcs $\gets []$}
    \For{$i \in 1,\dots,n$}
        \State $\text{pend}[i].id \gets i$
        \State $\text{pend}[i].e_l \gets RNN_L.init().append(v_i)$
        \State $\text{pend}[i].e_r \gets RNN_R.init().append(v_i)$
    \EndFor

    \While{$\left\vert{\text{pend}}\right\vert > 1$}
    \State{$A \gets \{ (i,d) \mid 1 \leq i < |\text{pend}|, d \in \{l,r\} \}$ }
    \State{$i,d \gets select(A)$}
    \If{$d = l$} \label{line:start}
        \State{$m,h \gets \text{pend}[i],\text{pend}[i+1]$}
        \State{$m.c  = m.e_l \circ m.e_r$}
        \State{$m.enc = g(W(m.c)+b)$}
        \State{$h.e_l.append(m.enc)$}
        \State{pend.remove($i$)}
    \Else
        \State{$h,m \gets \text{pend}[i],\text{pend}[i+1]$}
        \State{$m.c = m.e_l \circ m.e_r$}
        \State{$m.enc = g(W(m.c)+b)$}
        \State {$h.e_r.append(m.enc)$}
        \State{pend.remove($i+1$)}
    \EndIf \label{line:end}
    \State{arcs.add($h.id,m.id$)}
    \EndWhile{}
    \State{\Return arcs}
\end{algorithmic}
\caption{Parsing while maintaining tree representations}\label{alg:parsing2}
\end{algorithm}

\subsection{Labeled Tree Representation}

The tree representation described above does not account for the relation labels
$\ell$ the parsing algorithm assigns each edge. In cases the tree is fully specified in advance, the relation of each word to its head can be added to the word representations $v_i$.  However, in the context of parsing, the labels become known only when the modifier is attached to its parent.  We thus 
extend the tree representation by concatenating the node vector representation with a vector representation assigned to the label connecting the sub-tree to its parent. Formally, only the final $enc(t)$ equation changes:
\begin{align*}
enc(t) = g(W^e \cdot (e_l \circ e_r \circ \ell) + b^e)
\end{align*}

\noindent where $\ell$ is a learned embedding vector associated with the given label.

\subsection{Scoring Function}
\label{sec:scoring}

The parsing algorithm relies on a function $select(A)$ for choosing the
action to take at each stage.
We model this function as:
\begin{align*}
    select(A) = argmax_{(i,d,\ell) \in A}Score(pend,i,d,\ell)
\end{align*}

\noindent where $Score(.)$ is a learned function whose job is to assign scores
to possible actions to reflect their quality.  Ideally, it will not only score
correct actions above incorrect ones, but also more confident (easier) actions
above less confident ones, in order to minimize error propagation in the greedy
parsing process.

When scoring a possible attachment between a head $h$ and a modifier $m$ with
relation $\ell$, the scoring function should attempt to reflect the
following pieces of information:
\begin{itemize}
    \item Are the head words of $h$ and $m$ compatible under relation $l$?
    \item Is the modifier $m$ compatible with the already existing modifiers of
        $h$?  In other words, is $m$ a good subtree to connect as an outer-most
        modifier in the subtree $h$?
    \item Is $m$ complete, in the sense that it already acquired all of its own
        modifiers?
\end{itemize}
to this end, the scoring function looks at a window of $k$ subtrees to each side
of the head-modifier pair ($pend[i-k],\dots,pend[i+1+k]$)
where the neighbouring subtrees are used
for providing hints regarding possible additional modifiers of $m$ and $h$ that
are yet to be acquired.  We use $k=2$ in our experiments, for a total of 6
subtrees in total. This window approach is also used in the
Easy-First parser of Goldberg and Elhadad \cite{goldberg2010efficient} and works that extend it
\cite{tratz2011easyfirst,ma2012easyfirst,ma2013easyfirst}.
However, unlike the previous work, which made use of extensive feature
engineering and rich feature functions aiming at extracting the many relevant
linguistic sub-structures from the 6 subtrees and their interactions, we provide
the scoring function solely with the vector-encoding of the 6 subtrees in the
window.

Modeling the labeled attachment score is more difficult than modeling the
unlabeled score and is prone to more errors. Moreover, picking the label for an attachment will
cause less cascading error in contrast to picking the wrong attachment, which
will necessarily preclude the parser from reaching the correct tree structure.
In order to partially overcome this issue, our scoring function is a
sum of two auxiliary scoring function, one scoring unlabeled and the
other scoring labeled attachments. The unlabeled attachment score term in the sum
functions as a fallback which makes it easier for a parser to predict the attachment
direction even when there is no sufficient certainty as to the label.

\begin{align*}
Score(pend,i,d,\ell) &= Score_U(pend,i,d) \\
                  &+ Score_L(pend,i,d,\ell)
\end{align*}
\noindent Each of $Score_U$ and $Score_L$ are modeled as multi-layer perceptrons:
\begin{align*}
    &Score_U(pend,i,d) = MLP_U(x_i)[d] \\
    &Score_L(pend,i,d,\ell) = MLP_L(x_i)[(d,\ell)] \\
    &x_i = pend[i-2].c \circ\dots\circ pend[i+3].c 
\end{align*}
\noindent where $MLP_U$ and $MLP_L$
are standard multi-layer perceptron classifiers with one hidden layer
    ($MLP_X(x) = W^2g(W^1x+b^1)+b^2$)
and have output layers with size $2$ and
$2L$ respectively, $[.]$ is an indexing operation, and we assume the
values of $d$ and $(d,\ell)$ are mapped to integer values.

\subsection{Computational Complexity}

The Easy-First parsing algorithm works in $O(n\log n)$ time
\cite{goldberg2010efficient}.  The parser in this works differ by three aspects:
running a BI-LSTM encoder prior to parsing ($O(n)$); maintaining the tree
representation during parsing (lines \ref{line:start}--\ref{line:end} in
Algorithm \ref{alg:parsing2}) which take a constant time at each parsing step; and local
scoring using an MLP rather than a linear classifier (again, a constant-time
operation).  Thus, the parser maintains the $O(n\log n)$ complexity of the
Easy-First parser.

\section{Training Algorithm}
\label{sec:training}

\subsection{Loss and Parameter Updates}

At each step of the parsing process we select the highest scoring action
$(i,d,\ell)$. The
goal of training is to set the Score function such that correct actions are scored
above incorrect ones.  We use a margin-based objective, aiming to
maximize the margin between the highest scoring correct action and the set of
incorrect actions.
Formally, we define a hinge loss for each parsing step as
follows:
\begin{align*}
max\{ 0, 1 -& max_{(i,d,\ell) \in G}Score(pend,i,d,\ell)  \\
           +& max_{(i',d',\ell') \in A\setminus G}Score(pend,i,d,\ell)  \}
\end{align*}
\noindent where $A$ is the set of all possible actions and $G$ is the set of correct
actions at the current stage.

As the scoring function depends on vector-encodings of the all the trees in the
window, and each tree-encoding depends on the network's parameters, each parameter
update will invalidate all the vector encodings, requiring a re-computation of
the entire network.  We thus sum the local losses throughout the parsing
process, and update the parameter with respect to the sum of the losses at
sentence boundaries. Since we are using hinge loss the gradients will become sparser as the
training progresses. Fewer non-zero gradient could translate to unreliable updates.
In order to increase gradient stability and training speed,
we use a variation of mini-batch in which we update the parameters only after 50 errors were made.
This assures us a sufficient number of gradients for every update thus minimizing the effect of
gradient instability. 
The gradients of the entire network with respect to the sum of the losses are
calculated using the backpropagation algorithm.
Initial experiments with an SGD optimizer showed very
instable results.  We settled instead on using the ADAM optimizer
\cite{kingma2014adam} which
worked well without requiring fiddling with learning rates.

\vspace{-3pt}
\subsection{Error-Exploration and Dynamic Oracle Training}
\vspace{-3pt}
At each stage in the training process, the parser assigns scores to all the
possible actions $(i,d,\ell) \in A$. It then selects an action, applies it, and moves to the
next step.  Which action should be chosen?  A sensible option is to define $G$
as the set of actions that can lead to the gold tree, and following the highest
scoring actions in this set.
However, using training in this manner tends to suffer from error propagation at
test time.  The parser sees only states that result from following correct
actions. The lack of examples containing errors in the training phase makes
it hard for the parser to infer the best action given partly erroneous trees.
In order to cope with this, we follow the \emph{error exploration} training
strategy, in which we let the parser follow the highest scoring action in $A$
during training even if this action is incorrect, exposing it to states that
result from erroneous decisions.
This strategy requires defining the set $G$ such that the correct actions to
take are
well-defined also for states that cannot lead to the gold tree.
Such a
set $G$ is called a \emph{dynamic oracle}. 
Error-exploration and
dynamic-oracles were introduced by Goldberg and Nivre
\shortcite{coling2012dynamic}.
\paragraph{The Dynamic Oracle} A
dynamic-oracle for the easy-first parsing system we use is presented
in \cite{tacl2013dynamic}. 
Briefly, the dynamic-oracle version of $G$ defines the set of gold actions as 
the set of actions which does not increase the number of erroneous attachments
more than the minimum possible (given previous erroneous actions).
The number of erroneous attachments is increased in three cases:
(1) connecting a modifier to its head prematurely. Once the modifier is attached it is 
removed from the pending list and therefore can no longer acquire any of its own
modifiers;
(2) connecting a modifier to an erroneous head, when the correct head is still
on the pending list;
(3) connecting a modifier to a correct head, but an incorrect label.

Dealing with cases (2) and (3) is trivial.
To deal with (1), we consider as correct only actions in which the modifier is 
\emph{complete}. To efficiently identify complete modifiers
we hold a counter for
each word which is initialized to the number of modifiers the word has in the 
gold tree. When applying an attachment the counter of the modifier's gold head
word is decreased.
When the counter reaches 0, the sub-tree 
rooted at that word has no pending modifiers, and is considered complete.

\paragraph{Aggressive Exploration} We found that even when using error-exploration, after one iteration the model
remembers the training set quite well, and does not make enough errors to make
error-exploration effective.  In order to expose the parser to more errors, we 
employ a cost augmentation scheme: we sometimes follow incorrect actions also if
they score below correct actions.  Specifically, when the score of the correct
action is greater than that of the wrong action but the difference is smaller
than the margin constant, we chose to follow the wrong action with probability
$p_{aug}$ (we use $p_{aug}=0.1$ in our experiments).
Pseudocode for the entire training algorithm is given in the supplementary
material.

\subsection{Out-of-vocabulary items and word-dropout}

Due to the sparsity of natural language, we are likely to encounter at test
time a substantial number of the words that did not appear in the training data
(OOV words).  OOV words are likely even when pre-training the word
representations on a large un-annotated corpora.
A common approach is
to designate a special ``unknown-word'' symbol, whose associated vector will be used as the word
representation whenever an OOV word is encountered at test time.
In order to train the unknown-word vector, a possible approach is 
to replace all the words appearing in the training corpus less than a certain
number of times with the unknown-word symbol. This approach give a good vector
representation for unknown words but in the expense of ignoring many of the words from the training corpus.

We instead propose a variant of the word-dropout approach \cite{iyyer2015deep}.
During training, we replace a word with the unknown-word symbol with probability
that is inversely proportional to frequency of the word. Formally, we replace a
word $w$ appearing $\#(w)$ times in the training corpus with the unknown symbol
with a probability:
\vspace{-5pt}
\[
p_{unk}(w) = \frac{\alpha}{\#(w)+\alpha}
\]
\vspace{-5pt}

Using this approach we learn a vector representation for unknown words with minimal impact on the training of sparse words.

\section{Implementation Details}

Our Python implementation will be made available at the first author's website.
We use the PyCNN wrapper of the
CNN library\footnote{\url{https://github.com/clab/cnn/tree/master/pycnn}} for building the computation graph of the
network, computing the gradients using automatic differentiation, and performing
parameter updates.
We noticed the error on the development set does not improve after 20 iterations
over the training set, therefore, 
we ran the training for 20 iterations. The sentences where shuffled between
iterations. Non-projective sentences were skipped during training. We use the
default parameters initialization, step sizes and regularization values provided
by the PyCNN toolkit.
The hyper-parameters of the final networks used for all the reported experiments
are detailed in Table \ref{tbl:hyper}.

\begin{table}[ht]
\begin{center}
\begin{scalebox}{0.8}{
\begin{tabular}{ | c | c | }
\hline 
Word embedding dimension & 100 \\
POS tag embedding dimension & 25 \\
Relation embedding dimension & 25 \\
Hidden units in $Score_U$ & 100 \\
Hidden units in $Score_L$ & 100 \\
LSTM Dimensions (tree) & 200 \\
LSTM Layers (tree) & 2 \\
BI-LSTM Dimensions  & 100+100 \\
BI-LSTM Layers & 2 \\
Mini-batch size & 50 \\
$\alpha$ (for word dropout) & 0.25 \\
$p_{aug}$ (for exploration training) & 0.1 \\
$g$ & $tanh$ \\
\hline
\end{tabular}}
\end{scalebox}
\caption{Hyper-parameter values used in experiments}
\label{tbl:hyper}
\end{center}
\end{table}

Weiss et al \shortcite{weiss2015structured}
stress the importance of careful hyperparameter tuning for achieving top
accuracy in neural network based parser.
We did not follow this advice
and made very few attempts at hyper-parameter tuning, using manual hill climbing
until something seemed to work with reasonable accuracy, and then sticking with
it for the rest of the experiments. 

\section{Experiments and Results}

We evaluated our parsing model to English and Chinese data.
For comparison purposes we followed the setup of
\cite{dyer2015transitionbased}.

\paragraph{Data}
For English, we used the Stanford Dependency (SD) \cite{sd} conversion of the Penn
Treebank \cite{penntb}, using the standard train/dev/test splits
with the same predicted POS-tags as used in
\cite{dyer2015transitionbased,chen2014fast}. This dataset contains a few non-projective trees.
Punctuation symbols are excluded from the evaluation.

For Chinese, we use the Penn Chinese Treebank 5.1 (CTB5), using the
train/test/dev splits of
\cite{tale-two-parsers,dyer2015transitionbased} 
with gold
part-of-speech tags, also following
\cite{dyer2015transitionbased,chen2014fast}.

When using external word embeddings, we also use the same data as
\cite{dyer2015transitionbased}.\footnote{We thank Dyer et al for
sharing their data with us.}

\paragraph{Experimental configurations}
We evaluated the parser in several configuration. 
\textsc{BottomUpParser} is the
baseline parser, not using the tree-encoding, and instead representing each item
in $pending$ solely by the vector-representation (word and POS) of its head word.
\textsc{BottomUpParser+HtLstm} is using our Hierarchical Tree LSTM
representation. \textsc{BottomUpParser+HtLstm+bi-lstm} is the Hierarchical Tree
LSTM where we additionally use a BI-LSTM encoding for the head words.  Finally, we added
external, pre-trained word embeddings to the \textsc{BottomUpParser+HtLstm+bi-lstm} setup.
We also evaluated the final parsers in a \textsc{--pos} setup, in which we did
not feed the parser with any POS-tags.

\paragraph{Results}
Results for English and Chinese are presented in Tables \ref{res:en} and
\ref{res:cn} respectively.  For comparison, we also show the results of the
Stack-LSTM transition-based parser 
model of Dyer et al \shortcite{dyer2015transitionbased}, which we consider to be
a state-of-the-art greedy model which is also very competitive with search-based
models, with and without
pre-trained embeddings, and with and without POS-tags.
\begin{table}[t]
    \scalebox{0.8}{
\begin{tabular}{l|c c|c c}
& Dev && Test \\
& UAS & LAS & UAS & LAS \\
\hline
\textsc{BottomUpParser} & 83.3 & 79.0 & 82.7 & 78.6 \\
+\textsc{HtLstm} & 92.4 & 90.1 & 92.0 & 89.8 \\
$\;\;$+\textsc{bi-lstm} input & 93.0 & 90.5 & 92.6 & 90.2 \\
$\;\;\;\;$+external embeddings & 93.3 & 90.8 & 93.0 & 90.9 \\
\hline
\hline
Dyer et al (2015) no external & 92.7 & 90.4 & 92.4 & 90.0 \\
Dyer et al (2015) w/ external & 93.2 & 90.9 & 93.1 & 90.9 \\
C\&M (2014) w/ external & 92.2 & 89.7 & 91.8 & 89.6 \\
\hline
\hline
\textsc{BottomUp+All--pos} & 92.9 & 90.5 & 92.9 & 90.6 \\
Dyer et al (2015) --\textsc{pos} & 93.1 & 90.4 & 92.7 & 90.3 \\
\hline
\end{tabular}}
\caption{English parsing results (SD)}
\label{res:en}
\end{table}

\begin{table}[t]
    \scalebox{0.8}{
\begin{tabular}{l|c c|c c}
& Dev && Test \\
& UAS & LAS & UAS & LAS \\
\hline
\textsc{BottomUpParser} & 79.3 & 77.1 & 78.8 & 76.3 \\
+\textsc{HtLstm} & 86.2 & 84.5 & 86.2 & 84.7 \\
$\;\;$+\textsc{bi-lstm} & 86.2 & 84.5 & 86.1 & 84.4 \\
$\;\;\;\;$+external embeddings & 87.2 & 85.7 & 87.1 & 85.5 \\

\hline
\hline
Dyer et al (2015) no external & 86.3 & 84.7 & 85.7 & 84.1 \\
Dyer et al (2015) w/ external & 87.2 & 85.9 & 87.2 & 85.7 \\
C\&M (2014) no external & 84.0 & 82.4 & 83.9 & 82.4 \\
\hline
\hline
\textsc{BottomUp+All} \textsc{--pos} & 82.9 & 80.0 & 82.6 & 79.5 \\
Dyer et al (2015) \textsc{--POS} & 82.8 & 79.8 & 82.2 & 79.1 \\
\hline
\end{tabular}}
\caption{Chinese parsing results (CTB5)}
\label{res:cn}
\end{table}

The trends are consistent across the two languages.
The baseline Bottom-Up parser performs very poorly. This is expected, as only
the head-word of each subtree is used for prediction.  When adding the
tree-encoding, results jump to near state-of-the-art accuracy, suggesting that
the composed vector representation is indeed successful in capturing predictive structural
information.
Replacing the head-words with their BI-LSTM encodings results in another increase
in accuracy for English, outperforming the Dyer et al (\textsc{s-lstm} no external) models
on the test-set.  Adding the external pre-trained embeddings further improve the
results for both our parser and Dyer et al's model, closing the gap between
them.  
When POS-tags are not provided as input, the numbers for both parsers drop.
The drop is small for English and large for Chinese, and our parser seem to
suffer a little less than the Dyer et al model.

\paragraph{Importance of the dynamic oracle}
We also evaluate the importance of using the dynamic oracle and
error-exploration training, and find that they are indeed
important for achieving high parsing accuracies with our model (Table
\ref{tbl:dyn}). 

\begin{table}[h!]
    \centering
    \scalebox{0.8}{
    \begin{tabular}{l|cc|cc}
        & \multicolumn{2}{c}{English} &\multicolumn{2}{c}{Chinese} \\
        & UAS & LAS & UAS & LAS \\
        \hline
        \textsc{Rand}      & 93.0 & 90.5 & 86.2 & 84.5 \\
        \textsc{Rand-NoDyn}& 92.2 & 89.8 & 85.7 & 84.1 \\
        \hline
        \textsc{Ext}      & 93.3 & 90.8 & 87.2 & 85.7 \\
        \textsc{Ext-NoDyn}& 92.7 & 90.4 & 86.6 & 85.1 \\
        \hline
    \end{tabular}}
    \caption{Effect of the error-exploration training (dynamic-oracle) on dev
    set accuracy in English and Chinese. \textsc{Rand}: random
    initialization. \textsc{Ext}: pre-trained external embeddings.}
    \label{tbl:dyn}
\end{table}

When training without error-exploration (that is, the parser follows
only correct actions during training and not using the dynamic aspect of the oracle),
accuracies of unseen sentences drop by between 0.4 and 0.8 accuracy points
(average 0.58). This is consistent with previous work on training with
error-exploration and dynamic oracles \cite{goldberg2013training}, showing that
the technique is not restricted to models trained with sparse linear models.

\paragraph{Comparison to other state-of-the-art parsers}
Our main point of comparison is the model of Dyer et al, which was chosen because it is (a) a very strong parsing model; and (b) is the closest to ours in the literature: a greedy parsing
 model making heavy use of LSTMs.  To this end, we tried to make the comparison to Dyer et al as
controlled as possible, using the same dependency annotation schemes, as well as the same predicted POS-tags and the pre-trained embeddings (when applicable).

It is also informative to position our results with respect to other state-of-the-art parsing results reported in the literature, as we do in Table \ref{tbl:sota}.
Here, some of the comparisons are less direct: some of the results use different dependency annotation schemes\footnote{Our English parsing experiments use the Stanford Dependencies scheme, while other work use less informative dependency relations which are based on the Penn2Malt converter, using the Yamada and Matsumoto head rules. From our experience, this conversion is somewhat easier to parse, resulting in numbers which are about 0.3-0.4 points higher than Stanford Dependencies.}, as well as different predicted POS-tags, and different pre-trained word embeddings. While the numbers are not directly comparable, they do give a good reference as to the expected range of state-of-the-art parsing results. Our system's English parsing results are in range of state-of-the-art and the Chinese parsing results surpass it.  These numbers are achieved while using a greedy, bottom up parsing method without any search, and while relying solely on the compositional tree representations. 
\begin{table*}
    \centering
\scalebox{0.8}{
\begin{tabular}{l|c|c|c|c|c|c|c}
System & Method & Representation & Emb & PTB-YM & PTB-SD & CTB & Runtime \\
\hline
ZhangNivre11 & transition (beam) & large feature set (sparse) & -- & 92.9 & -- &
86.0 & $O(n)+$ \\
Martins13 & graph, 3rd order+ & large feature set (sparse) & -- & 92.8 & 93.07 & -- & $O(n^4)$ \\ 
Pei15 & graph, 2nd order & large feature set (dense) & -- & 92.99 & -- & -- & $O(n^3)$ \\
This Work & EasyFirst (greedy) & Rec-LSTM encoding & -- &  -- & 92.6 &  86.1 & $O(n\log n)$ \\
\hline
Weiss15 & transition (greedy) & large feature set (dense) & YES & -- & 93.19 & -- & $O(n)$  \\
Weiss15 & transition (beam) & large feature set (dense) & YES & -- & 93.99 & -- & $O(n)+$ \\
Pei15 & graph, 2nd order & large feature set (dense) & YES & 93.29 & -- & -- & $O(n^3)$ \\
LeZuidema14 & reranking /blend & inside-outside recursive net & YES & 93.12 & 93.84 & -- & $O(n^3)$\\
Zhu15 & reranking /blend & recursive conv-net & YES & 93.83 & -- & 85.71 & $O(n)+$ \\
This Work & EasyFirst (greedy) & Rec-LSTM encoding & YES &  -- & 93.0 &  87.1 & $O(n\log n)$\\
\end{tabular}}
\caption{\footnotesize Parsing results (UAS) of various state-of-the-art parsing systems on the English and Chinese datasets. The systems that use embeddings use different pre-trained embeddings.
English results use predicted POS tags (different systems use different
taggers), while Chinese results use gold POS tags. 
\textbf{PTB-YM}: English PTB, Yamada and Matsumoto
head rules. \textbf{PTB-SD}: English PTB, Stanford Dependencies (different
systems may use different versions of the Stanford converter. \textbf{CTB}:
Chinese Treebank.  
\emph{reranking /blend} in method column indicates a reranking system where the
reranker score is interpolated with the base-parser's score. The reranking
systems' runtimes are those of the base parsers they use. $O(n)+$
indicates a linear-time system with a large multiplicative constant.
The different systems and the numbers reported from them are taken from:
ZhangNivre11: \cite{zhang11acl}; Martins13: \cite{martins2013turbo}; Weiss15
\cite{weiss2015structured}; Pei15: \cite{pei2015effective}; LeZuidema14
\cite{le2014insideoutside}; Zhu15: \cite{zhu2015reranking}. }
\label{tbl:sota}
\end{table*}

\vspace{-5pt}
\section{Related Work}
\label{sec:related}
\vspace{-5pt}

We survey two lines of related work: methods for encoding trees as vectors, and
methods for parsing with vector representations.

The popular approach for encoding trees as vectors is using recursive neural
networks \cite{goller1996learning,socher2010learning,tai2015improved}. Recursive neural networks represent the
vector of a parent node in a tree as a function of its children nodes. However,
the functions are usually restricted to having a fixed maximum arity (usually
two) \cite{socher2010learning,tai2015improved,socher2014recursive}.
While trees can be binarized to cope with the
arity restriction, doing so result in deep trees which in turn lead to the
vanishing gradient problem when training. To cope with the vanishing gradients,
\cite{tai2015improved} enrich the composition function with a gating mechanism
similar to that of the LSTM, resulting in the so-called Tree-LSTM model.
Another approach is to allow arbitrary arities but ignoring the sequential
nature of the modifiers, e.g. by using a bag-of-modifiers representation or a
convolutional layer \cite{tai2015improved,zhu2015reranking}.  
In contrast, our tree encoding method naturally allows for arbitrary branching
trees by relying on the well established LSTM sequence model, and using it as a
black box.  
Very recently, \cite{lapata} proposed an RNN-based tree encoding which is
similar to ours in encoding the sequence of modifiers as an RNN. Unlike our
bottom-up encoder, their method works top-down, and is therefor not readily
applicable for parsing.  On the other hand the top-down approach is well suited for
generation.  In future work, it could be interesting to combine the bottom-up
and top-down approaches in an encoder-decoder framework
\cite{seq2seq,skipthought}.
Work by Dyer et al \shortcite{dyer2016recurrent}, that was
submitted in parallel to ours, introduces a similar LSTM-based representation of
syntactic constituents in the context of phrase-grammar parsing.

In terms of parsing with vector representations, there are four dominant
approaches: search based parsers that use local features that are fed to a
neural-network classifier \cite{pei2015effective,durrett2015neural}; greedy transition based parsers
that use local features that are fed into a neural-network classifier
\cite{chen2014fast,weiss2015structured}, sometimes coupled with a node composition function
\cite{dyer2015transitionbased,watanabe2015transitionbased}; bottom up parsers that rely solely on recursively combined
vector encodings of subtrees
\cite{socher2010learning,stenetorp2013transition,socher2013parsing}; and parse-reranking
approaches that first produced a $k$-best list of parses using a
traditional parsing technique, and then scores the trees based on a recursive
vector encoding of each node \cite{le2014insideoutside,le2015forest,zhu2015reranking}.

Our parser is a greedy, bottom up parser that rely on compositional
vector encodings of subtrees as its sole set of features.  Unlike the re-ranking
approaches, we do not rely on an external parser to provide $k$-best lists. Unlike
the bottom-up parser in \cite{socher2010learning} who only parses sentences of up
to 15 words and the parser of \cite{stenetorp2013transition} who achieves
very low parsing accuracies, we parse arbitrary sentences with near state-of-the-art
accuracy. Unlike the bottom up parser in
\cite{socher2013parsing} we do not make use of a grammar. 
The parser of \cite{weiss2015structured} obtains exceptionally high results using local
features and no composition function.  
The greedy version
of their parser 
uses
extensive tuning of hyper-parameters and network depth in order to squeeze every possible bit of
accuracy. Adding beam search on top of that further improves results. Due to our much more limited resources, we did not perform a
methodological search over hyper-parameters, and explored only a tiny space of
the possible hyper-parameters, and our parser does not perform search.  
Finally, perhaps closest to our
approach is the greedy, transition-based parser of \cite{dyer2015transitionbased} that also
works in a bottom-up fashion, and incorporates an LSTM encoding of the input
tokens and hierarchical vector composition into its scoring mechanism.  Indeed,
that parser obtains similar scores to ours, although we obtain somewhat better
results when not using pre-trained embeddings.  We differ from the parser of
Dyer et al by having a more elaborate vector-composition function, relying solely on
the compositional representations, and
performing fully bottom-up parsing without being guided by a stack-and-buffer
control structure.

\vspace{-5pt}
\section{Conclusions and Future Work}
\vspace{-5pt}
We suggest a compositional vector representation of parse trees that
relies on a recursive combination of recurrent-neural network encoders, and
demonstrate its effectiveness by integrating it in a bottom-up easy-first
parser.  Future extensions in terms of parsing include the addition of beam
search, handling of unknown-words using character-embeddings, and
adapting the algorithm to constituency trees.  We also plan to establish the
effectiveness of our Hierarchical Tree-LSTM encoder by applying it to more semantic
vector representation tasks, i.e. training tree representation for capturing
sentiment \cite{socher2013recursive,tai2015improved},
semantic sentence similarity \cite{sick} or textual inference \cite{snli}.

\paragraph{Acknowledgements}
This research is supported by the Intel Collaborative Research Institute for
Computational Intelligence (ICRI-CI) and the Israeli Science Foundation
(grant number 1555/15).


\bibliographystyle{acl}

\clearpage

\setcounter{algorithm}{2}
\onecolumn
\section*{Appendix: Training Algorithm Pseudocode}
\begin{algorithm*}[h!]
\begin{algorithmic}[1]
\State{\textbf{Input:} Sentences $w^1,\dots,w^m$}
\State{\textbf{Input:} Tree annotations $T^1,\dots,T^m$}
\State{\textbf{Input:} Number of epochs to train}

\
\State{$V \gets InitializeVectors()$}
\State{$Loss \gets []$}

\
\For{$epoch \in \{1,\dots,Epochs\}$}
	\For{$S,T \in \{(w^1,T^1),\dots,(w^m,T^m)\}$}
    	\State{$Loss \gets TrainSentence\left(S, V[w_1,\dots,w_n], T, Loss\right)$}

      \
      \If{$\left\vert{Loss}\right\vert > 50$}
      \State{$SumLoss \gets sum(Loss)$}
      \State{$Call\ ADAM\ to\ minimize\ SumLoss$}
      \State $Loss \gets []$
      \EndIf
	\EndFor
\EndFor
\end{algorithmic}
\caption{Training on annotated corpus}\label{alg:trainalg-full}
\end{algorithm*}

\noindent(See Algorithm 4, training of a single sentence, on next page.)

\begin{algorithm*}[h!]
\begin{algorithmic}[1]
\Function{TrainSentence}{$w,v,T, Loss$}
    \State{\textbf{Input:} Sentence $w = w_1,\dots,w_n$}
    \State{\textbf{Input:} Vectors $v_i$ corresponding to inputs $w_i$}
    \State{\textbf{Input:} Annotated tree $T$ in the form of $(h,m,rel)$ triplets}
    \State{\textbf{Input:} List $Loss$ to which loss expressions are added}
    
    \
    \For{$i \in 1,\dots,n$}
		\State $unassigned[i] \gets \left\vert{Children(w_i)}\right\vert$
        \State $pend[i].id \gets i$
        \State $pend[i].e_l \gets RNN_L.init().append(v_i)$
        \State $pend[i].e_r \gets RNN_R.init().append(v_i)$
    \EndFor

	\
    \While{$\left\vert{pend}\right\vert > 1$}
    \State{$G, W \gets \left\{\right\}, \left\{\right\} $}
   
   \
    \For{$(i,d,rel) \in \{ 1 \leq i < |pend|, d \in \{l,r\}, rel \in Relations \}$ }
    \If{$d = l$}{\ $m,h \gets \text{pend}[i],\text{pend}[i+1]$}
    \Else{\ $m,h \gets \text{pend}[i+1],\text{pend}[i]$}
    \EndIf
    
    \
    \If{$unassigned[m.id] \neq 0 \lor \exists_{\ell \neq rel}{ (h,m,\ell) \in T}$}
    	\State{$W.append((h,m,rel))$}
    \Else{\ $G.append((h,m,rel))$}
    \EndIf{}
    \EndFor{}
    
    \
    \State{$h_G,m_G,rel_G \gets argmax_{(i,d,\ell) \in G}Score(pend,i,d,\ell)$}
    \State{$h_W,m_W,rel_W \gets argmax_{(i,d,\ell) \in W}Score(pend,i,d,\ell)$}
    \State{$score_G \gets Score(h_G,m_G,rel_G)$}
    \State{$score_W \gets Score(h_W,m_W,rel_W)$}
    
    \
    \If{$score_G - score_W < 0$}
    \State{$h,m,rel,score \gets h_W,m_W,rel_W,score_W$}
    \ElsIf{$score_G - score_W > 1 \lor random()<p_{aug}$}
    \State{$h,m,rel,score \gets h_G,m_G,rel_G,score_G$}
    \Else
    \State{$h,m,rel,score \gets h_W,m_W,rel_W,score_W$}
    \EndIf

    \If{$score_G - score < 1$}
    \State{$Loss.append(1 - score_G + score)$}
    \EndIf
    
    \
    \State{$m.c  = m.e_l \circ m.e_r$}
    \State{$m.enc = g(W(m.c \circ rel)+b)$}
    \If{$h.id < m.id$}{\ $h.e_l.append(m.enc)$}
    \Else{\ $h.e_r.append(m.enc)$}
    \EndIf

	\
	\State{$unassigned[T_{Parent}(m).id] \gets unassigned[T_{Parent}(m).id] - 1$}
    \State{pend.remove($m$)}
    \EndWhile{}    
    \EndFunction
\State{\Return Loss}
\caption{Training on a single sentence with dynamic oracle algorithm}

\end{algorithmic}
\end{algorithm*}

\end{document}